\documentclass{article}
\usepackage{lmodern}

\PassOptionsToPackage{numbers, compress}{natbib}
\usepackage[preprint]{neurips_2025}

\usepackage[utf8]{inputenc} % allow utf-8 input
\usepackage[T1]{fontenc}    % use 8-bit T1 fonts
\usepackage{url}            % simple URL typesetting
\usepackage{booktabs}       % professional-quality tables
\usepackage{amsfonts}       % blackboard math symbols
\usepackage{nicefrac}       % compact symbols for 1/2, etc.
\usepackage{microtype}      % microtypography
\usepackage{xcolor}         % colors

\usepackage{multirow}   % Needed for multirow table cells
\usepackage{float}
\usepackage{amsmath}
\usepackage{booktabs}   % For better looking tables
\usepackage{graphicx}

\title{Are Data Embeddings effective in time series forecasting?}

\author{%
  Reza Nematirad \\
  Kansas State University \\
  \texttt{nematirad@ksu.edu}  % Replace with actual email
  \And
  Anil Pahwa \\
  Kansas State University \\
  \texttt{pahwa@ksu.edu}  % Replace with actual email
  \And
  Balasubramaniam Natarajan \\
  Kansas State University \\
  \texttt{bala@ksu.edu}  % Replace with actual email
}

\begin{document}

\maketitle

\begin{abstract}
Time series forecasting plays a crucial role in many real-world applications, and numerous complex forecasting models have been proposed in recent years. Despite their architectural innovations, most state-of-the-art models report only marginal improvements—typically just a few thousandths in standard error metrics. These models often incorporate complex data embedding layers to transform raw inputs into higher-dimensional representations to enhance accuracy. But are data embedding techniques actually effective in time series forecasting? Through extensive ablation studies across fifteen state-of-the-art models and four benchmark datasets, we find that removing data embedding layers from many state-of-the-art models does not degrade forecasting performance—in many cases, it improves both accuracy and computational efficiency. The gains from removing embedding layers often exceed the performance differences typically reported between competing state-of-the-art models. Code is available at \url{https://github.com/neuripsdataembedidng/DataEmbedding}.

\end{abstract}

\section{Introduction}
\label{gen_inst}
Time series forecasting is a fundamental task in machine learning with broad applications, including energy systems, traffic management, healthcare, finance, and weather prediction \cite{wang2024timexer}. In recent years, numerous deep learning frameworks have been proposed to improve forecasting performance. These models often employ complex architectures such as statistical components, Transformers, Multilayer perceptrons (MLPs), and Convolutional neural networks (CNNs). Despite architectural diversity, recent state-of-the-art models achieve gains of only a few thousandths of a point in common metrics such as mean squared error (MSE) or mean absolute error (MAE). Table \ref{Table1} presents the performance of several time series forecasting models \cite{nematirad2025times2d, yu2025lino, liu2023itransformer, li2024revisiting, wang2023micn, wang2024timemixer, han2024softs, dai2024periodicity} that report state-of-the-art results. Despite substantial research efforts, the actual performance gains from these advanced models remain modest.
%%%%%%%%%%%%%%%%%%%%%%%%%%%%% Table 1 %%%%%%%%%%%%%%%%%%%%%%%%%%%%%%%%%%
\begin{table}[H]
\caption{
Performance of recent multivariate forecasting models for prediction horizon $H \in \{96, 192, 336, 720\}$ and input length $L = 96$. 
\textbf{\textcolor{red}{Red}} and \textcolor{blue}{blue} denote best and second-best results.
}
\centering
\resizebox{\textwidth}{!}{%
\begin{tabular}{cccccccccccccccccc}
\toprule
\multicolumn{2}{c}{\textbf{Models}} & 
\multicolumn{2}{c}{Times2D} & 
\multicolumn{2}{c}{LiNo} & 
\multicolumn{2}{c}{iTransformer} & 
\multicolumn{2}{c}{RLinear} &  
\multicolumn{2}{c}{MICN} & 
\multicolumn{2}{c}{TimeMixer} & 
\multicolumn{2}{c}{SOFTS} & 
\multicolumn{2}{c}{PDF} \\
\multicolumn{2}{c}{} & 
\multicolumn{2}{c}{(2025)} & 
\multicolumn{2}{c}{(2025)} & 
\multicolumn{2}{c}{(2024)} & 
\multicolumn{2}{c}{(2024)} & 
\multicolumn{2}{c}{(2024)} & 
\multicolumn{2}{c}{(2024)} & 
\multicolumn{2}{c}{(2024)} & 
\multicolumn{2}{c}{(2024)} \\
\textbf{Data} & H & MSE & MAE & MSE & MAE & MSE & MAE & MSE & MAE & MSE & MAE & MSE & MAE & MSE & MAE & MSE & MAE \\
\cmidrule(lr){1-1}  % Under "Data"
\cmidrule(lr){2-2}  % Under "H"
\cmidrule(lr){3-4}
\cmidrule(lr){5-6}
\cmidrule(lr){7-8}
\cmidrule(lr){9-10}
\cmidrule(lr){11-12}
\cmidrule(lr){13-14}
\cmidrule(lr){15-16}
\cmidrule(lr){17-18}

\multirow{4}{*}{\rotatebox{90}{ETTh1}} 
& 96  & \textbf{\textcolor{red}{0.378}} & \textbf{\textcolor{red}{0.394}} & \textcolor{blue}{0.379} & \textcolor{blue}{0.395} & 0.386 & 0.405 & 0.395 & 0.419 & 0.404 & 0.428 & 0.375 & 0.400 & 0.381 & 0.399 & 0.387 & 0.405 \\
& 192 & 0.431 & \textcolor{blue}{0.422} & \textbf{\textcolor{red}{0.423}} & 0.423 & 0.441 & 0.436 & \textcolor{blue}{0.424} & 0.445 & 0.471 & 0.471 & 0.429 & \textbf{\textcolor{red}{0.421}} & 0.435 & 0.431 & 0.439 & 0.438 \\
& 336 & 0.463 & \textbf{\textcolor{red}{0.436}} & \textcolor{blue}{0.455} & \textcolor{blue}{0.438} & 0.487 & 0.458 & \textbf{\textcolor{red}{0.446}} & 0.466 & 0.571 & 0.538 & 0.484 & 0.458 & 0.480 & 0.452 & 0.494 & 0.464 \\
& 720 & \textcolor{blue}{0.473} & 0.464 & \textbf{\textcolor{red}{0.459}} & \textcolor{blue}{0.456} & 0.503 & 0.491 & 0.470 & 0.488 & 0.651 & 0.622 & 0.498 & 0.482 & 0.499 & \textbf{\textcolor{red}{0.448}} & 0.491 & 0.484 \\
\midrule

\multirow{4}{*}{\rotatebox{90}{ETTm1}} 
& 96  & \textcolor{blue}{0.324} & 0.363 & 0.322 & \textcolor{blue}{0.361} & 0.334 & 0.368 & 0.329 & 0.367 & \textbf{\textcolor{red}{0.320}} & 0.374 & \textbf{\textcolor{red}{0.320}} & \textbf{\textcolor{red}{0.357}} & 0.325 & 0.361 & 0.335 & 0.367 \\
& 192 & 0.370 & 0.386 & \textcolor{blue}{0.365} & \textcolor{blue}{0.383} & 0.377 & 0.391 & 0.367 & 0.385 & 0.378 & 0.414 & \textbf{\textcolor{red}{0.361}} & \textbf{\textcolor{red}{0.381}} & 0.375 & 0.389 & 0.377 & 0.393 \\
& 336 & 0.402 & \textcolor{blue}{0.406} & 0.401 & 0.408 & 0.426 & 0.420 & \textcolor{blue}{0.399} & 0.410 & 0.428 & 0.452 & \textbf{\textcolor{red}{0.390}} & \textbf{\textcolor{red}{0.404}} & 0.405 & 0.412 & 0.408 & 0.415 \\
& 720 & 0.459 & \textbf{\textcolor{red}{0.439}} & 0.469 & 0.447 & 0.491 & 0.459 & \textcolor{blue}{0.454} & 0.483 & 0.482 & \textbf{\textcolor{red}{0.441}} & \textcolor{blue}{0.454} & \textbf{\textcolor{red}{0.441}} & 0.466 & 0.447 & 0.457 & 0.448 \\

\bottomrule
\end{tabular}
}
\label{Table1}
\end{table}

%%%%%%%%%%%%%%%%%%%%%%%%%% End of Table 1 %%%%%%%%%%%%55
Furthermore, time series forecasting algorithms consist of complex and advanced components. However, their effectiveness and their individual contributions to overall forecasting performance are not adequately investigated. One prominent component is data embedding, which transforms raw input data into higher-dimensional representations. For instance, PDF \cite{dai2024periodicity} and Times2D \cite{nematirad2025times2d} apply various data embedding techniques without sufficiently justifying the rationale behind using them. On the other hand, models such as PatchTST \cite{nie2023a}, SOFTS \cite{han2024softs}, MICN \cite{wang2023micn}, and ETSFormer \cite{woo2023etsformer} provide specific justifications for incorporating particular embedding techniques into their architecture. However, the effectiveness of the utilized embedding techniques is not adequately discussed. Consequently, it is unclear whether data embedding techniques truly improve forecasting performance.

Motivated by the minimal improvements achieved through increasingly complex models (Table \ref{Table1}) and insufficient evaluation of core components, we revisit the effectiveness of data embedding layers in time series forecasting. We directly investigate a simple yet important question: \textbf{are data embeddings actually effective in time series forecasting?}

Our claim is simple but promising: \textbf{removing the data embedding layers from many state-of-the-art forecasting models does not degrade forecasting performance—in many cases, it enhances both forecasting accuracy and computational efficiency. Interestingly, the gains from removing embedding layers often exceed the performance differences typically reported between competing state-of-the-art models}. Our goal is not to imply that data embedding will never be effective in time series forecasting. Instead, we aim to highlight our promising findings and suggest that the community should devote greater attention to critically assessing the actual impact of embedding layers in existing models.

We substantiate our claims by conducting extensive experiments using fifteen time series forecasting models on four standard benchmark datasets originally reported in their studies. Each selected model explicitly utilizes data embedding as a core architectural component. First, we make a great effort to reproduce the results of the standard time series models, using their publicly available publications and the code provided in their official repositories. Next, we identify the data embedding components in each model and rerun the models with these embedding layers bypassed. It should be noted that, in the absence of embedding layers, some preprocessing steps, such as permutation and concatenation are performed to reconcile the input with the model expected dimensions. Then, we evaluate forecasting performance and computational efficiency, including epoch runtime and memory usage. The contributions of this study are summarized as follows:
\begin{itemize}
\item To our knowledge, this is the first systematic study to rigorously evaluate the effectiveness of data embedding layers in time series forecasting models.
\item We conduct comprehensive ablation studies on fifteen high-performing forecasting algorithms across standard benchmark datasets. We show that removing embedding layers generally does not degrade forecasting performance—in many cases, it enhances both forecasting accuracy and computational efficiency.
\item We highlight that the gains from removing embedding layers often exceed the performance differences typically reported between competing state-of-the-art models. This finding emphasizes the importance of carefully evaluating the model components before adding further complexity.
\end{itemize}

\section{Data Embedding}
\label{others}
Data embedding refers to techniques that transform raw input data into structured representations \cite{koshil2024towards}. These transformations are widely used in modern time series forecasting models to prepare data for downstream neural network components. Data embedding strategies can be categorized as follows. A summarized comparison of the most common embedding techniques is provided in Table~\ref{tab:scientific_embedding_categories} in the Appendix.

\subsection{Value Embedding}
Value embedding refers to the transformation of raw input time series into a latent feature space, typically of higher dimension. Formally, given a multivariate input sequence $\mathbf{X} \in \mathbb{R}^{B \times L \times C}$, where $B$ is the batch size, $L$ is the sequence length, and $C$ is the number of input channels (features), a value embedding module projects $\mathbf{X}$ into an embedding space of dimension $d$ by mapping the channel dimension $C \rightarrow d$. Two value embedding methods are commonly used \cite{li2024lagcnn}:
\begin{itemize}
\item \textbf{Token-based convolutional embedding}: Applies a 1D convolution along the temporal axis to project input features into a higher-dimensional space. This operation can be expressed as 
\[
\mathbf{Z} = \text{Conv1D}(\mathbf{X})
\]

where $\mathbf{Z} \in \mathbb{R}^{B \times L \times d}$.
\item \textbf{Linear projection}: Applies a linear transformation independently at each time step and maps each input feature vector $\mathbf{x}_t \in \mathbb{R}^C$ to an embedding vector in $\mathbb{R}^d$.
\end{itemize}

\subsection{Temporal embedding}
Temporal embedding encodes time-related features such as minute, hour, day, or month into continuous vectors. Two main types of temporal embedding are used in forecasting models \cite{li2021multivariate}:
\begin{itemize}
\item \textbf{Discrete temporal embedding}: This approach embeds categorical time fields (e.g., hour of day, day of week, month) using one of the following techniques: \textbf{Fixed embedding}: Uses non-trainable sinusoidal vectors to map each discrete time index to a fixed vector based on sine and cosine functions. \textbf{Learnable embedding}: Implements trainable lookup tables (via nn.Embedding) that map each discrete temporal category to a vector learned during training. 

\item \textbf{Continuous time feature embedding}: This method encodes normalized continuous features (e.g., scaled hour or day values) using a linear projection. Time features are fed into a fully connected layer that maps them to the embedding space $\mathbb{R}^d$.
\end{itemize}

\subsection{Positional Embedding}

Positional embedding injects information about the position of each time step in the sequence, which is not inherently modeled by components like attention or MLPs. Unlike value or temporal embeddings, which depend on the content of the input features or time-related fields, positional embeddings are purely based on the position index in the sequence. Given a model embedding dimension \( d \) and maximum sequence length \( L \), the positional embedding constructs a fixed matrix \( \mathbf{P} \in \mathbb{R}^{1 \times L \times d} \), where each time index \( t \in [0, L) \) is mapped to a deterministic vector using sine and cosine functions at varying frequencies:
\[
\mathbf{P}_{t, 2i} = \sin\left( \frac{t}{10000^{2i/d}} \right), \quad
\mathbf{P}_{t, 2i+1} = \cos\left( \frac{t}{10000^{2i/d}} \right)
\]
Here, \( i \in \left[ 0, \left\lfloor \frac{d}{2} \right\rfloor \right) \) denotes the embedding dimension index. The constant \( 10000 \) is an empirically chosen scaling factor that defines the base of the frequency spectrum and ensures smooth variation across dimensions. The resulting positional embedding tensor is broadcast across the batch dimension and typically added to the value and/or temporal embeddings before being passed to the model \cite{chen2023contiformer}.

\subsection{Inverted Embedding}

Inverted embedding refers to a design where both the raw input features and the associated time-based features (e.g., hour, day, month) are concatenated along the feature dimension. Given an input time series $\mathbf{X} \in \mathbb{R}^{B \times L \times C}$ and its corresponding time covariates $\mathbf{X}_{\text{mark}} \in \mathbb{R}^{B \times L \times C_{\text{time}}}$, the two tensors are concatenated across the last axis to form:
\[
\mathbf{X}_{\text{concat}} = \text{Concat}(\mathbf{X}, \mathbf{X}_{\text{mark}}) \in \mathbb{R}^{B \times L \times (C + C_{\text{time}})}
\]

The resulting tensor is then transposed to shape $\mathbb{R}^{B \times (C + C_{\text{time}}) \times L}$. A linear projection is then applied to map the sequence dimension $L$ into the model embedding space of dimension $d$:
\[
\mathbf{Z} = \text{Linear}(\mathbf{X}_{\text{concat}}^\top), \quad \mathbf{Z} \in \mathbb{R}^{B \times (C + C_{\text{time}}) \times d}
\]

Unlike traditional approaches that treat each \textit{time step} as a token, inverted embedding treats each \textit{variable} as a token. In value, temporal, and positional embeddings, the model maps or enriches the feature dimension, keeping the sequence length fixed. In contrast, inverted embedding applies the projection over the sequence length, allowing the model to focus on the temporal patterns of each variable \cite{han2024softs, wan2025ppdformer}.

\subsection{Patch Embedding}
Patch embedding refers to a technique that segments the input time series into overlapping or non-overlapping patches. Each patch is then projected into a latent embedding space. This enables the model to process sequences at the granularity of temporal segments. As a result, it can capture localized patterns and reduce input length for long sequences.

Given a multivariate time series input $\mathbf{X} \in \mathbb{R}^{B \times L \times C}$, patching is applied along the temporal axis. The sequence for each variable is divided into fixed-length patches of size $P$, using a sliding window with stride $S$. This results in a new sequence of patches of length $\lfloor \frac{L - P}{S} + 1 \rfloor$, each of shape $\mathbb{R}^{P}$.
Each patch is then projected into an embedding dimension $d$ using a linear layer:
\[
\mathbf{z}_{i} = \mathbf{W} \mathbf{x}_{i}, \quad \mathbf{W} \in \mathbb{R}^{d \times P}
\]
where $\mathbf{x}_{i} \in \mathbb{R}^{P}$ is the $i$-th patch and $\mathbf{z}_{i} \in \mathbb{R}^{d}$ is its corresponding embedding \cite{nie2023a, dai2024periodicity}.
%%%%%%%%%%%%%%%%%%%%%%%%%%%%%%%%%%%%%%%%%%%%%%%%%%%%%%%%%%%%%%%%%%%%%%%%%%%%%%%%%%%%
\section{Related Works}
%%%%%%%%%%%%%%%%%%%%%%%% Data Embedding ######################
The Multi-scale Isometric Convolution Network (MICN) \cite{wang2023micn} decomposes the input into seasonal and trend components. A value embedding via 1D convolution (token embedding) is applied to the seasonal sequence, combined with a time-based embedding using fixed values and a sinusoidal positional embedding. These three embedded components are summed and passed through a dropout layer before being fed into the model.
ETSformer \cite{woo2023etsformer} proposes a Transformer architecture inspired by exponential smoothing, using decomposed components for level, growth, and seasonality. It employs a value embedding module implemented via 1D convolution to map input features into a latent space. 
WITRAN \cite{jia2023witran} introduces a bi-granular recurrent framework for time series forecasting that models short- and long-term repetitive patterns through 2D information flows. The model concatenates raw input features and time-based covariates along the feature dimension. Then, a fixed temporal embedding is applied to the feature dimension. 

%%%%%%%%%%%%%%%%%%%%%%%% Inverted ######################
The Series-cOre Fused Time Series forecaster (SOFTS) \cite{han2024softs} is an efficient MLP-based framework that introduces the STar Aggregate-Redistribute (STAR) module, which employs a centralized strategy to aggregate all series into a global core representation. SOFTS employs an inverted embedding mechanism. It uses linear projection to combine multivariate feature values and time-related metadata. Then, the combined representation is mapped into a high-dimensional space over the sequence dimension. 
EDformer \cite{chakraborty2024edformerembeddeddecompositiontransformer} introduces a decomposition-based Transformer that separates multivariate time series into trend and seasonal components. It adopts an inverted embedding strategy by concatenating the seasonal component with time-based features and projecting the sequence dimension into a latent space using a linear layer.
PPDformer \cite{wan2025ppdformer} also adopts an inverted embedding design. It concatenates denoised multivariate feature values with time-based features across the feature axis, and maps the sequence into the embedding space using a linear projection.

Times2D \cite{nematirad2025times2d} proposes a multi-block decomposition that transforms raw multivariate time series into 2D periodic segments using the Fast Fourier Transform. These segments are passed through 2D convolutional layers, followed by flattening to produce embeddings. This patch-style embedding does not rely on common value, temporal, or positional embeddings.

Crossformer \cite{zhang2023crossformer}introduces a hierarchical Transformer architecture that models both temporal and cross-variable dependencies for multivariate forecasting. It employs a dual-embedding mechanism. First, it employs a patch-based embedding strategy where the multivariate time series is first segmented into fixed-length patches using a sliding window. Then, each patch is projected into a latent space using the summation of value embedding through linear layers and sinusoidal positional embeddings within the patches.

PatchTST \cite{nie2023a} proposes a channel-independent Transformer for time series forecasting, where each univariate time series is processed separately. It applies the same patch-based embedding mechanism. Unlike Crossformer, PatchTST focuses solely on modeling temporal dependencies within each variable and does not capture cross-variable interactions.

\section{Experimental Setup}
\paragraph{Baselines and Datasets.}We evaluate fifteen high-performing time series forecasting models. These models have been introduced in top-tier venues in artificial intelligence and machine learning. Models selected in this study cover a broad spectrum of architectural paradigms. Transformer-based architectures include Crossformer \cite{zhang2023crossformer}, PatchTST \cite{nie2023a}, ETSformer \cite{woo2023etsformer}, iFlowformer, and iFlashAttention \cite{kang2025vardrop}. MLP-based approaches comprise MICN \cite{wang2023micn}, SOFTS \cite{han2024softs}, EDformer \cite{chakraborty2024edformerembeddeddecompositiontransformer}, LiNo \cite{yu2025lino}, Minusformer \cite{liang2024minusformer}, and VarDrop \cite{kang2025vardrop}. Finally, hybrid and decomposition-based frameworks are represented by Times2D \cite{nematirad2025times2d}, PDF \cite{dai2024period}, PPDformer \cite{wan2025ppdformer}, and WITRAN \cite{jia2023witran}.
We evaluate all models on four widely-used benchmark datasets: ETTh1, ETTh2, ETTm1, and ETTm2. These datasets consist of multivariate time series that capture diverse temporal patterns across various operational domains. ETTh1 and ETTh2 are recorded at hourly intervals, while ETTm1 and ETTm2 are sampled every 15 minutes \cite{jin2024timellm}. Additional details on the datasets are provided in Table~\ref{table_dataset_summary} in the Appendix.

\paragraph{Setup and Evaluation Metric.} 
All input time series are normalized using the mean and standard deviation from the training set. The input sequence length is fixed at 96, and prediction horizons are set to \( H \in \{96, 192, 336, 720\} \), consistent across all models and datasets. Forecasting accuracy is evaluated using MSE and MAE, while computational efficiency is measured by the average training time per epoch (in seconds) and memory usage (in MB).

\paragraph{Infrastructure.} \label{sec_infra}
All simulations are conducted on a high-performance Linux workstation equipped with an NVIDIA L4 GPU (23 GB memory), CUDA version 12.4, and dual AMD EPYC 7713 64-core processors (128 threads in total). The system has 1 TB of RAM and runs on Ubuntu with Python 3.10 and PyTorch 2.2.1.

\section{Results}
We present comprehensive results comparing the performance of each model with and without data embedding layers. The forecasting performance results are summarized in Tables~\ref{tab:etth1}, \ref{tab:etth2}, \ref{tab:ettm1}, and \ref{tab:ettm2}, corresponding to the ETTh1, ETTh2, ETTm1, and ETTm2 datasets, respectively. Below, we highlight key trends observed across models and datasets.

%%%%%%%%%%%%%%%%%%%%%%%%%%%%%%%%%%%%%%%%% Table ETTh1 #######################################################
\begin{table}[htbp]
\caption{ETTh1 forecasting results with and without embedding for input length \( L = 96 \) and prediction horizons \( H \in \{96, 192, 336, 720\} \). \textbf{Bold} values indicate better performance.}
  \label{tab:etth1}
  \centering
  \resizebox{\textwidth}{!}{%
\begin{tabular}{llcccccccccccc}
    \toprule
    Model & Metric & \multicolumn{6}{c}{With Embedding} & \multicolumn{6}{c}{Without Embedding} \\
\cmidrule(lr){3-8} \cmidrule(lr){9-14}
                   &        & \multicolumn{4}{c}{H} & Time & Mem & \multicolumn{4}{c}{H} & Time & Mem \\
\cmidrule(lr){3-8} \cmidrule(lr){9-14}
                   &        & 96 & 192 & 336 & 720 &        &        & 96 & 192 & 336 & 720 &        &        \\
    \midrule
    PDF    & MSE & 0.387   & 0.439  & 0.494  & 0.491  & 48.01  & 2857       & \textbf{0.377}   & \textbf{430}  & \textbf{0.484}  & \textbf{0.503}  & \textbf{16.50}  & \textbf{2760}  \\
                & MAE & 0.405   & 0.438  & 0.464  & 0.484  &       &            & \textbf{0.401}   & \textbf{0.429}  & \textbf{0.453}  & \textbf{0.481}  &        &   \\

\midrule
    ETSformer   & MSE & 0.564   & 0.747  & 0.987  & 0.987  & 24.61  & 4506       & \textbf{0.563}   & \textbf{0.611}  & \textbf{0.643}  & \textbf{0.627}  & \textbf{10.95}  & \textbf{4496}  \\
                & MAE & 0.536   & 0.651  & 0.788  & 0.806  &       &            & \textbf{0.505}   & \textbf{0.528}  & \textbf{0.545}  & \textbf{0.558}  &        &   \\

\midrule
    PatchTST    & MSE & 0.389   & 0.449  & 0.498  & 0.544  & 10.90  & 4351       & \textbf{0.385}   & \textbf{0.438}  & \textbf{0.488}  & \textbf{0.541}           & \textbf{8.41}  & 4353  \\
                & MAE & 0.409   & 0.445  & 0.474  & 0.517  &       &            & \textbf{0.404}   & \textbf{0.433}  & \textbf{0.459}  & \textbf{0.511}  &        &   \\

\midrule

    MICN        & MSE & 0.404   & 0.471  & 0.576  & 0.651  & 19.75  & 2739       & \textbf{0.402}  & \textbf{0.450}  & \textbf{0.475}  & \textbf{0.531}  & \textbf{5.52}  & \textbf{2709}  \\
                & MAE & 0.428   & 0.471  & 0.538  & 0.622  &       &             & \textbf{0.421}   & \textbf{0.448}  & \textbf{0.473}  & \textbf{0.527}  &        &   \\

\midrule
    SOFTS      & MSE & 0.385   & 0.445  & 0.501  & 0.565  & 9.93  & 2706       & \textbf{0.383}  & \textbf{0.444}  & \textbf{0.486}  & \textbf{0.519}  & \textbf{7.97}  & \textbf{2223}  \\
                & MAE & 0.405   & 0.441  & 0.469  & 0.529  &       &            & \textbf{0.401}  & \textbf{0.439}  & \textbf{0.462}  & \textbf{0.502}  &        &   \\

\midrule
    VarDrop     & MSE & 0.416   & 0.447  & 0.490  & 0.537  & 9.94  & 497       & \textbf{0.386}   & \textbf{0.442}  & \textbf{0.491}  & \textbf{0.495}  & \textbf{8.37}  & \textbf{395}  \\
                & MAE & 0.425   & 0.445  & 0.466  & 0.504  &       &            & \textbf{0.408}   & \textbf{0.439}  & \textbf{0.467}  & \textbf {0.488}  &        &   \\
\midrule
    Crossformer & MSE & \textbf{0.390}  & 0.561 & 0.639 & 0.921 & 40.48 & 4396      & 0.404 & \textbf{0.501} & \textbf{0.634} & \textbf{0.871} & \textbf{40.32} & \textbf{4383} \\
                & MAE & \textbf{0.421}  & 0.543 & 0.588 & 0.755 &     &            & 0.427 & \textbf{0.493} & \textbf{0.581} & \textbf{0.739} &      &  \\

\midrule
    iFlashAttention       & MSE & 0.407   & 0.456  & 0.487   & 0.5532  & 10.80  & 2293       & \textbf{0.387}   & \textbf{0.443}   & 0.490           & \textbf{0.492}  & \textbf{9.75}  &  \textbf{2292}  \\
                          & MAE & 0.420   & 0.451  & 0.467   & 0.5131  &       &             & \textbf{0.407}   & \textbf{0.439}   & \textbf{0.467}  & \textbf{0.486}  &        &   \\                         
\midrule
    iFlowformer       & MSE & 0.394   & 0.459  & 0.493  & 0.545  & 12.40  & 2297       & \textbf{0.391}   & \textbf{0.441}  & \textbf{0.479}  & \textbf{0.499}  & \textbf{07.69}  & \textbf{2281}  \\
                      & MAE & 0.408   & 0.450  & 0.466  & 0.508  &       &             & \textbf{0.409}   & \textbf{0.440}  & \textbf{0.458}  & \textbf{0.490}  &        &   \\
\midrule
    PPDformer   & MSE & 0.415   & 0.460  & 0.496  & 0.506  & 36.08 & 2738       & \textbf{0.398}   & 0.470  & \textbf{0.473}  & \textbf{0.487}  & \textbf{17.92}  & \textbf{2709}  \\
                & MAE & 0.424   & 0.451  & 0.468  & 0.492  &       &            & \textbf{0.419}   & 0.455  & \textbf{0.461}  & \textbf{0.486}  &        &   \\

\midrule
    LiNo    & MSE & 0.379   & 0.443  & 0.476  & 0.496  & 3.83  & 2036       & \textbf{0.372}   & \textbf{0.429}  & \textbf{0.454}  & \textbf{0.460}  & \textbf{3.61}  & \textbf{2026}  \\
            & MAE & 0.395   & 0.432  & 0.446  & 0.474  &       &            & \textbf{0.389}   & \textbf{0.427}  & \textbf{0.436}  & \textbf{0.458}  &        &   \\

\midrule
    EDformer    & MSE & 0.433   & 0.520  & 0.582  & \textbf{0.661}  & \textbf{3.52}  & \textbf{2260}       & \textbf{0.420}   & \textbf{0.493}  & \textbf{0.546}  & 0.666  & 3.64  & 2661  \\
                & MAE & 0.449   & 0.504  & 0.537  & \textbf{0.608}  &       &                              & \textbf{0.441}   & \textbf{0.488}  & \textbf{0.519}  & 0.618  &        &   \\

\midrule
    Minusformer   & MSE & 0.382   & 0.431  & 0.481  & 0.522  & 8.03  & \textbf{2395}       & \textbf{0.374}   & \textbf{0.425}  & \textbf{0.477}  & \textbf{0.520}  & \textbf{7.44}  & 2693  \\
                  & MAE & 0.398   & 0.430  & 0.454  & \textbf{0.492}  &       &            & \textbf{0.395}   & \textbf{0.429}  & \textbf{0.450}  & 0.493  &        &   \\

\midrule
    WITRAN   & MSE & 0.552   & 0.646  & 0.757  & 0.899  & \textbf{16.34}  & 2050       & \textbf{0.545}   & \textbf{0.634}  & \textbf{0.764}  & \textbf{0.895}  & 16.38  & \textbf{2036}  \\
             & MAE & 0.548   & 0.608  & 0.676  & 0.746  &       &                     & \textbf{0.541}   & \textbf{0.599}  & \textbf{0.659}  & \textbf{0.746}  &        &   \\

\midrule
    Times2D  & MSE & 0.378   & 0.431  & 0.463  & 0.473  & \textbf{5.58}  & 778       & \textbf{0.359}   & \textbf{0.427}  & \textbf{0.461}  & \textbf{0.472}  & 6.48  & \textbf{758}  \\
             & MAE & 0.394   & 0.422  & 0.436  & 0.464  &       &                    & \textbf{0.383}   & \textbf{0.421}  & \textbf{0.435}  & \textbf{0.463}  &        &   \\

\bottomrule
\end{tabular}
}
\end{table}

%%%%%%%%%%%%%%%%%%%%%%%%%%%%%%%%%%%%%%%%%%%%%%%%%%%%% Table ETTh2 %%%%%%%%%%%%%%%%%%%%%%%%%%%%%%%%
\begin{table}[htbp]
\caption{ETTh2 forecasting results with and without embedding for input length \( L = 96 \) and prediction horizons \( H \in \{96, 192, 336, 720\} \). \textbf{Bold} values indicate better performance.}
  \label{tab:etth2}
  \centering
  \resizebox{\textwidth}{!}{%
  \begin{tabular}{llcccccccccccc}
    \toprule
    Model & Metric & \multicolumn{6}{c}{With Embedding} & \multicolumn{6}{c}{Without Embedding} \\
\cmidrule(lr){3-8} \cmidrule(lr){9-14}
                   &        & \multicolumn{4}{c}{H} & Time & Mem & \multicolumn{4}{c}{H} & Time & Mem \\
\cmidrule(lr){3-8} \cmidrule(lr){9-14}
                   &        & 96 & 192 & 336 & 720 &        &        & 96 & 192 & 336 & 720 &        &        \\
\midrule
    PDF       & MSE & 0.307   & 0.376  & 0.414  & 0.437  & 47.82 & 2838        & \textbf{0.300}   & \textbf{0.375}  & \textbf{0.412}  & \textbf{0.431}  & \textbf{16.45}  & \textbf{2752}  \\
              & MAE & 0.353   & 0.401  & 0.426  & 0.452  &       &             & \textbf{0.348}   & \textbf{0.397}  & \textbf{0.426}  & \textbf{0.447}  &        &   \\    
    
\midrule
    ETSformer   & MSE & 0.399   & 0.521  & 0.615  & 0.692  & 23.50  & 2721      & \textbf{0.345}   & \textbf{0.436}  & \textbf{0.487}  & \textbf{0.494}  & \textbf{8.80}  & \textbf{2080}  \\
                & MAE & 0.435   & 0.505  & 0.569  & 0.616  &       &            & \textbf{0.399}   & \textbf{0.446}  & \textbf{0.483}  &\textbf{0.497}  &        &   \\

\midrule
    PatchTST    & MSE & 0.300   & 0.382  & 0.435  & 0.447  & 14.96  & 2351      & \textbf{0.297}   & \textbf{0.375}  & \textbf {0.413}  & \textbf {0.418}  & \textbf {6.46}  & \textbf {2758}  \\
                & MAE & 0.350   & 0.404  & 0.441  & 0.463  &       &            & \textbf{0.384}   & \textbf{0.399}  & \textbf {0.428}  & \textbf {0.440}  &        &   \\

\midrule
    MICN        & MSE & 0.354   & 0.475  & 0.602  & 0.829  & 17.95  & 2761       & \textbf{0.333}   & \textbf{0.463}  & \textbf{0.567}  & \textbf{0.804}  & \textbf{5.24}  & \textbf{2708}  \\
                & MAE & 0.400   & 0.477  & 0.540  & 0.654  &       &             & \textbf{0.389}   & \textbf{0.472}  & \textbf{0.527}  & \textbf{0.646}  &        &   \\

\midrule
    SOFTS   & MSE & 0.305   & 0.375  & 0.437  & 0.439  & 10.06  & 634       & \textbf{0.295}   & \textbf{0.374}  & \textbf{0.416}  & \textbf{0.434}  & \textbf{8.09}  & \textbf{473}  \\
            & MAE & 0.350   & 0.396  & 0.437  & \textbf{0.447}  &       &            & \textbf{0.347}   & \textbf{0.395}  & \textbf{0.430}  & 0.450  &        &   \\

\midrule
    VarDrop     & MSE & 0.306    & 0.393  & 0.423  & 0.436  & 9.642  & 2288       & \textbf{0.303}   & \textbf{0.383}  & \textbf{0.423}  & \textbf{0.418}  & \textbf{8.39}  & \textbf{2274}  \\
                & MAE & 0.355    & 0.406  & 0.438  & 0.452  &       &            & \textbf{0.353}   & \textbf{0.400}  & \textbf{0.436}  & \textbf{0.442}  &        &   \\
    \midrule
    Crossformer & MSE & 0.588   & 0.978  & 0.996  & 1.161  & 56.51  & 2243       & \textbf{0.573}   & \textbf{0.757}  & \textbf{0.796}  & \textbf{0.945}  & \textbf{56.41}  & \textbf{2223}  \\
                & MAE & 0.576   & 0.698  & 0.709  & 0.787  &       &            & \textbf{0.537}   & \textbf{0.628}  & \textbf{0.643}  & \textbf{0.803}  &        &   \\

\midrule
    iFlashAttention      & MSE & 0.306   & 0.392  & 0.428  & 0.442  & 11.43  & 2303        & \textbf{0.305}   & \textbf{0.382}  & \textbf{0.426}  & \textbf{0.417}  & \textbf{10.00}  & \textbf{2294}  \\
                         & MAE & 0.355   & 0.406  & 0.438  & 0.455  &       &              & \textbf{0.353}   & \textbf{0.400}  & \textbf{0.430}  & \textbf{0.440}  &        &   \\

\midrule
    iFlowformer       & MSE & \underline{0.308}   & 0.389  & 0.431  & 0.440  & 13.02  & 2293       & \underline{0.308}   & \textbf{0.382}  & \textbf{0.422}  & \textbf{0.432}  & \textbf{10.70}  & \textbf{2281}  \\
                      & MAE & 0.357   & 0.409  & 0.440  & 0.453  &       &             & \textbf{0.355}   & \textbf{0.402}  & \textbf{0.433}  & \textbf{0.447}  &        &   \\

\midrule
    PPDformer       & MSE & 0.321   & 0.405  & 0.439  & 0.461  & 36.94  & 2727       & \textbf{0.320}   & \textbf{0.401}  & \textbf{0.436}  & \textbf{0.438}  & \textbf{22.24}  & \textbf{2714}  \\
                & MAE & 0.365   & 0.415  & 0.445  & 0.465  &       &             & \textbf{0.361}   & \textbf{0.406}  & \textbf{0.439}  & \textbf{0.454}  &        &   \\
                
\midrule
    LiNo   & MSE & 0.305   & 0.384  & 0.389  & 0.417  & 3.96  & 2035       & \textbf{0.296}   & \textbf{0.378}  & \textbf{0.385}  & \textbf{0.412}  & \textbf{3.83}  & \textbf{2021}  \\
           & MAE & 0.352   & 0.398  & 0.413  & 0.436  &       &            & \textbf{0.345}   & \textbf{0.395}  & \textbf{0.411}  & \textbf{0.432}  &        &   \\

\midrule
    EDformer    & MSE & 0.422   & 0.485  & 0.549  & 0.799  & \textbf{3.66}  & \textbf{2288}                & \textbf{0.404}   & \textbf{0.484}  & \textbf{0.504}  & \textbf{0.688}  & 3.98  & 2661  \\
                & MAE & 0.429   & 0.464  & 0.500  & 0.621  &                &                              & \textbf{0.423}   & \textbf{0.482}  & \textbf{0.490}  & \textbf{0.593}  &        &   \\

\midrule
    Minusformer  & MSE & 0.304   & 0.379  & 0.430  & 0.427  & 10.94  & 2420       & \textbf{0.293}   & \textbf{0.373}  & \textbf{0.421}  & \textbf{0.422}  & \textbf{8.78}  & \textbf{2699}  \\
                 & MAE & 0.349   & 0.396  & 0.435  & 0.441  &        &            & \textbf{0.344}   & \textbf{0.393}  & \textbf{0.429}  & \textbf{0.439}  &        &   \\

\midrule
    WITRAN   & MSE & 1.659   & 2.810  & 2.641  & 3.488  & 27.12  & 804       & 1.771   & 2.752  & 2.632  & 3.752  & 25.12  & 803  \\
             & MAE & 1.055   & 1.464  & 1.415  & 1.663  &       &            & 1.120   & 1.439  & 1.403  & 1.682  &        &   \\

\midrule
    Times2D   & MSE & \textbf{0.292}   & 0.376  & 0.379  & 0.413  & 5.52  & 775       & 0.294   & \textbf{0.371}  & \textbf{0.376}  & \textbf{0.406}  & \textbf{5.19}  & \textbf{740}  \\
              & MAE & \textbf{0.340}   & 0.391  & 0.407  & 0.434  &       &           & 0.342   & \textbf{0.390}  & \textbf{0.405}  & \textbf{0.429}  &        &   \\

\bottomrule
\end{tabular}
}
\end{table}

%%%%%%%%%%%%%%%%%%%%%%%%%%%%%%%%%%%%%%%%%%%%%%%%%%%%% Table ETTm1 %%%%%%%%%%%%%%%%%%%%%%%%%%%%%%%%
\begin{table}[htbp]
\caption{ETTm1 forecasting results with and without embedding for input length \( L = 96 \) and prediction horizons \( H \in \{96, 192, 336, 720\} \). \textbf{Bold} values indicate better performance.}
  \label{tab:ettm1}
  \centering
  \resizebox{\textwidth}{!}{%
  \begin{tabular}{llcccccccccccc}
    \toprule
    Model & Metric & \multicolumn{6}{c}{With Embedding} & \multicolumn{6}{c}{Without Embedding} \\
\cmidrule(lr){3-8} \cmidrule(lr){9-14}
                   &        & \multicolumn{4}{c}{H} & Time & Mem & \multicolumn{4}{c}{H} & Time & Mem \\
\cmidrule(lr){3-8} \cmidrule(lr){9-14}
                   &        & 96 & 192 & 336 & 720 &        &        & 96 & 192 & 336 & 720 &        &        \\

\midrule
    PDF     & MSE & 0.335   & 0.377  & 0.408  & 0.457  & 194.3  & 2961                       & \textbf{0.321}   & \textbf{0.365}  & \textbf{0.392}  & \textbf{0.451}  & \textbf{64.44}  & \textbf{2880}  \\
            & MAE & 0.367   & 0.393  & 0.4152  & 0.448  &       &                            & \textbf{0.359}   & \textbf{0.386}  & \textbf{0.405}  & \textbf{0.442}  &        &   \\
            
\midrule
    ETSformer   & MSE & 0.526   & 0.577  & 0.677  & 0.802  & 94.28  & 2803                      & \textbf{0.373}   & \textbf{0.408}  & \textbf{0.441}  & \textbf{0.499}  & \textbf{33.18}  & \textbf{2200}  \\
                & MAE & 0.515   & 0.553  & 0.620  & 0.708  &       &                            & \textbf{0.397}   & \textbf{0.410}  & \textbf{0.429}  & \textbf{0.462}  &        &   \\    

   \midrule
    PatchTST    & MSE & \textbf{0.344}   & 0.375  & 0.407  & 0.473  & 58.62  & 2476           & 0.348   & \textbf{0.370}  & \textbf{0.393}  & \textbf{0.459}  & \textbf{23.48}  & \textbf{2893}  \\
                & MAE & \textbf{0.367}   & 0.395  & 0.415  & 0.453  &       &                 & 0.371   & \textbf{0.387}  & \textbf{0.406}  & \textbf{0.440}  &        &   \\

\midrule
    MICN        & MSE & \textbf{0.320}   & 0.378  & 0.428  & 0.483  & 70.17  & 2850             & 0.354            & \textbf{0.363}   & \textbf{0.416}  & \textbf{0.478}  & \textbf{17.05}  & \textbf{2805}  \\
                & MAE & \textbf{0.374}   & 0.414  & 0.452  & 0.482  &       &                   & 0.380            & \textbf{0.395}  & \textbf{0.416}  & \textbf{0.455}  &        &   \\

\midrule
    SOFTS       & MSE & 0.325   & 0.384  & 0.429  & 0.477  & 33.51  & \textbf{2403}          & \textbf{0.223}   & \textbf{0.367}  & \textbf{0.407}  & \textbf{0.475}  & \textbf{27.16}  & 2419  \\
                & MAE & 0.361   & 0.397  & 0.423  & 0.455  &        &                        & \textbf{0.341}   & \textbf{0.386}  & \textbf{0.411}  & \textbf{0.452}  &        &   \\

\midrule
    VarDrop     & MSE & 0.340   & 0.398  & 0.439  & \textbf{0.490}  & 36.26  & 504     & 0.344   & \textbf{0.382}  & \textbf{0.428}  & 0.505          & \textbf{27.86}  & \textbf{393}  \\
                & MAE & 0.375   & 0.403  & 0.427  & \textbf{0.457}  &        &                            & 0.378   & \textbf{0.397}  & \textbf{0.425}  & 0.467  &        &   \\
                
\midrule
    Crossformer & MSE & 0.366   & 0.413  & 0.453  & 0.867  & 233.8  & 2237      & \textbf{0.363}   & \textbf{0.406}  & \textbf{0.447}  & \textbf{0.511}  & \textbf{230.4}  & \textbf{2228}  \\
                & MAE & 0.406   & 0.427  & 0.454  & 0.711  &       &            & \textbf{0.404}   & \textbf{0.418}  & \textbf{0.446}  & \textbf{0.481}  &        &   \\                
                
\midrule
    iFlashAttention  & MSE & 0.350   & 0.402  & 0.442  & 0.500  & 45.11  & 2438           & \textbf{0.344}   & \textbf{0.382}  & \textbf{0.434}  & \textbf{0.527}  & \textbf{34.28}  & \textbf{2421}  \\
                     & MAE & 0.381   & 0.405  & 0.428  & 0.464  &       &                 & \textbf{0.378}   & \textbf{0.397}  & \textbf{0.428}  & \textbf{0.476}  &        &   \\

\midrule
    iFlowformer       & MSE & 0.340   & 0.418  & \textbf{0.420}  & \textbf{0.492}  & 45.66  & 2409                & \textbf{0.339}   & \textbf{0.388}  & 0.448  & 0.501  & \textbf{34.97}  & \textbf{2331}  \\
                      & MAE & 0.373   & 0.412  & \textbf{0.424}  & \textbf{0.461}  &       &                      & \textbf{0.372}   & \textbf{0.399}  & 0.434  & 0.470  &        &   \\

\midrule
    PPDformer   & MSE & 0.356   & 0.411  & 0.440  & 0.503  & 149.5  & 2840       & \textbf{0.339}   & \textbf{0.389}  & \textbf{0.432}  & \textbf{0.493}  & \textbf{86.75}  & \textbf{2833}  \\
                & MAE & 0.392   & 0.420  & 0.438  & 0.471  &       &             & \textbf{0.376}   & \textbf{0.400}  & \textbf{0.429}  & \textbf{0.467}  &        &   \\

\midrule
    LiNo    & MSE & 0.331   & 0.400  & 0.435  & 0.503  & 13.50  & 2146                      & \textbf{0.323}   & \textbf{0.375}  & \textbf{0.418}  & \textbf{0.497}  & \textbf{12.47}  & \textbf{2146}  \\
            & MAE & 0.365   & 0.404  & 0.423  & 0.463  &       &                            & \textbf{0.361}   & \textbf{0.390}  & \textbf{0.416}  & \textbf{0.462}  &        &   \\

\midrule
    EDformer    & MSE & 0.395   & 0.432  & 0.486  & \textbf{0.544}  & \textbf{10.87}  & \textbf{2404}                      & \textbf{0.378}   & \textbf{0.426}  & \textbf{0.463}  & 0.551  & 11.36  & 2805  \\
                & MAE & 0.427   & 0.448  & 0.478  & \textbf{0.509}  &       &                            & \textbf{0.413}   & \textbf{0.443}  & \textbf{0.466}  & 0.520  &        &   \\

\midrule
    Minusformer  & MSE & 0.351   & 0.384  & \textbf{0.451}  & 0.491  & 40.98  & 2528              & \textbf{0.330}   & \textbf{0.381}  & 0.449          & \textbf{0.490}  & \textbf{32.57}  & \textbf{2847}  \\
                 & MAE & 0.377   & 0.394  & 0.432  & 0.459  &       &                             & \textbf{0.367}   & \textbf{0.392}  & \textbf{0.428}  & \textbf{0.457}  &        &   \\

\midrule
    WITRAN      & MSE & 0.640   & 0.769  & 0.827  & 0.951  & 107.8  & 937     & \textbf{0.637}   & \textbf{0.688}  & \textbf{0.803}  & \textbf{0.860}  & \textbf{107.6}  & \textbf{914}  \\
                & MAE & 0.590   & 0.668  & 0.714  & 0.766  &       &                  & \textbf{0.585}   & \textbf{0.622}  & \textbf{0.699}  & \textbf{0.713}  &        &   \\
                
\midrule
    Times2D     & MSE & 0.324   & 0.370  & 0.402  & 0.459  & 21.30  & 783           & \textbf{0.324}   & \textbf{0.368}  & \textbf{0.397}  & \textbf{0.458}  & \textbf{21.09}  & \textbf{761}  \\
                & MAE & 0.363   & 0.386  & 0.406  & 0.439  &       &               & \textbf{0.361}   & \textbf{0.383}  & \textbf{0.403}  & \textbf{0.438}  &        &   \\

\bottomrule
\end{tabular}
}
\end{table}

%%%%%%%%%%%%%%%%%%%%%%%%%%%%%%%%%%%%%%%%%%%%%%%%%%%%% Table ETTm2 
\begin{table}[htbp]
\caption{ETTm2 forecasting results with and without embedding for input length \( L = 96 \) and prediction horizons \( H \in \{96, 192, 336, 720\} \). \textbf{Bold} values indicate better performance.}
  \label{tab:ettm2}
  \centering
  \resizebox{\textwidth}{!}{%
  \begin{tabular}{llcccccccccccc}
    \toprule
    Model & Metric & \multicolumn{6}{c}{With Embedding} & \multicolumn{6}{c}{Without Embedding} \\
    \cmidrule(lr){3-8} \cmidrule(lr){9-14}
                   &        & \multicolumn{4}{c}{H} & Time & Mem & \multicolumn{4}{c}{H} & Time & Mem \\
    \cmidrule(lr){3-8} \cmidrule(lr){9-14}
                   &        & 96 & 192 & 336 & 720 &        &        & 96 & 192 & 336 & 720 &        &        \\

\midrule
    PDF     & MSE & 0.183   & 0.246  & 0.300  & 0.403  & 193.9 & 2975       & \textbf{0.176}   & \textbf{0.241}  & 0.304  & \textbf{0.402}  & \textbf{66.29}  & \textbf{2877}  \\
            & MAE & 0.265   & 0.307  & 0.342  & 0.401  &       &            & \textbf{0.258}   & \textbf{0.301}  & 0.344  & 0.403  &        &   \\

\midrule
    ETSformer   & MSE & 0.267   & 0.333  & 0.398  & 0.501  & 92.92  & 2806       & \textbf{0.189}   & \textbf{0.255}  & \textbf{0.318}  & \textbf{0.432}  & \textbf{32.23}  & \textbf{2209}  \\
                & MAE & 0.372   & 0.409  & 0.444  & 0.495  &       &             & \textbf{0.284}   & \textbf{0.323}  & \textbf{0.361}  & \textbf{0.427}  &        &   \\

\midrule
    PatchTST    & MSE & 0.183   & 0.248  & \textbf{0.311}  & \textbf{0.407}  & 58.71  & 2472       & \textbf{0.177}   & \textbf{0.244}  & 0.314  & 0.410  & \textbf{24.58}  & \textbf{2458}  \\
                & MAE & 0.263   & 0.309  & 0.351  & \textbf{0.402}  &       &             & \textbf{0.261}   & \textbf{0.305}  & \textbf{0.350}  & 0.408  &        &   \\
\midrule
    MICN    & MSE & \textbf{0.183}   & \textbf{0.272}  & 0.396  & 0.579  & 70.61  & 2845       & 0.193   & 0.280  & \textbf{0.314}   & \textbf{0.508}  & \textbf{17.24}  & \textbf{2836}  \\
            & MAE & \textbf{0.280}   & \textbf{0.346}  & 0.429  & 0.532  &       &             & 0.292   & 0.358  & \textbf{0.350}  & \textbf{0.498}  &        &   \\

\midrule
    SOFTS       & MSE & 0.180   & 0.251  & 0.315  & 0.417  & 30.73  & 638       &  \textbf{0.179}   &  \textbf{0.247}  &  \textbf{0.309}  &  \textbf{0.411}  &  \textbf{28.43}  &  \textbf{481}  \\
                & MAE & 0.262   & 0.309  & 0.349  & 0.407  &       &            &  \textbf{0.263}   &  \textbf{0.308}  &  \textbf{0.346}  &  \textbf{0.405}  &        &   \\   

\midrule
    VarDrop     & MSE & 0.182   & 0.250  & \textbf{0.312}  & \textbf{0.41}  & 36.48  & 499         & \textbf{0.181}   & \textbf{0.250}  & 0.322  & 0.421  & \textbf{28.34}  & \textbf{391}  \\
                & MAE & 0.266   & 0.311  & \textbf{0.350}  & \textbf{0.405}  &       &              & \textbf{0.265}   & \textbf{0.309}  & 0.357  & 0.410  &        &   \\

    \midrule
    Crossformer & MSE & \textbf{0.237}   & 0.450  & \textbf{0.640}  & 1.660  & 232.8  & 2145       & 0.257           & \textbf{0.437}  & 0.701  & \textbf{1.520}  & \textbf{232.7}  & \textbf{2131}  \\
                & MAE & 0.342           & 0.462  & \textbf{0.548}  & 0.914  &       &            & \textbf{0.336}   & \textbf{0.442}  & 0.602  & \textbf{0.886}  &        &   \\

\midrule
    iFlashAttention    & MSE & 0.182   & 0.250  & 0.312  & 0.411  & 39.93  & 2389       & 0.194   & \textbf{0.249}  & 0.322  & 0.416  & \textbf{33.69}  & \textbf{2375}  \\
                       & MAE & 0.266   & 0.311  & 0.349  & 0.405  &       &             & 0.279   & \textbf{0.309}  & 0.357  & 0.408  &        &   \\

\midrule
    iFlowformer   & MSE & 0.183   & 0.249  & 0.311  & \textbf{0.409}  & 32.70  & 2391       & \textbf{0.181}   & \textbf{0.248}  & \textbf{0.312}  & 0.420  & \textbf{27.12}  & \textbf{2391}  \\
                  & MAE & 0.269   & 0.310  & 0.349  & \textbf{0.404}  &       &             & \textbf{0.267}   & \textbf{0.307}  & \textbf{0.348}  & 0.410  &        &   \\

\midrule
    PPDformer   & MSE & 0.188   & 0.269  & 0.322  & 0.417  & 148.7  & 2870      & \textbf{0.180}   & \textbf{0.254}  & \textbf{0.308}  & \textbf{0.407}  & \textbf{86.57}  & \textbf{2802}  \\
                & MAE & 0.276   & 0.329  & 0.357  & 0.411  &       &            & \textbf{0.260}   & \textbf{0.307}  & \textbf{0.345}  & \textbf{0.403}  &        &   \\

\midrule
    LiNo    & MSE & 0.177   & 0.244  & 0.309  & 0.404  & 13.98  & \textbf{2147}       & \textbf{0.173}   & \textbf{0.241}  & \textbf{0.304}  & \textbf{0.403}  & \textbf{11.19}  & 2171  \\
            & MAE & 0.260   & 0.304  & 0.346  & \textbf{0.398}  &       &             & \textbf{0.256}   & \textbf{0.301}  & \textbf{0.342}  & 0.399  &        &   \\

\midrule
    EDformer    & MSE & 0.310   & 0.500  & \textbf{0.647}  & \textbf{0.755}  & 12.34  & \textbf{2397}      & \textbf{0.262}   & \textbf{0.449}  & 0.668  & 0.776  & \textbf{10.93}  & 2804  \\
                & MAE & 0.388   & 0.492  & \textbf{0.590}  & 0.637  &       &            & \textbf{0.353}   & \textbf{0.474}  & 0.607           & \textbf{0.617}  &        &   \\

\midrule
    Minusformer & MSE & 0.183   & 0.248  & 0.309  & 0.409  & 41.26  & 2543       & \textbf{0.176}   & \textbf{0.246}  & \textbf{0.308}  & \textbf{0.401}  & \textbf{33.69}  & \textbf{2848}  \\
                & MAE & 0.268   & 0.308  & 0.347  & 0.402  &       &             & \textbf{0.260}   & \textbf{0.304}  & \textbf{0.345}  & \textbf{0.400}  &        &   \\

\midrule
    WITRAN      & MSE & 0.807   & 1.136  & \textbf{1.293}  & 4.448  & 109.5  & 943       & \textbf{0.795}   & \textbf{1.092}  & 1.313  & \textbf{4.439}  & 107.5  & \textbf{925}  \\
                & MAE & 0.722   & 0.903  & \textbf{0.916}  & 1.793  &       &                     & \textbf{0.709}   & \textbf{0.886}  & 0.965  & \textbf{1.628}  &        &   \\  
                
\midrule
    Times2D     & MSE & 0.179   & 0.241  & 0.301  & 0.397  & \textbf{20.72}  & 781      & \textbf{0.175}   & \textbf{0.240}  & \textbf{0.300}  & \textbf{0.394}  & 21.53  & \textbf{764}  \\
                & MAE & 0.263   & 0.301  & 0.339  & 0.394  &       &                    & \textbf{0.256}   & \textbf{0.299}  & \textbf{0.338}  & \textbf{0.392}  &        &   \\

\bottomrule
\end{tabular}
}
\end{table}
%%%%%%%%%%%%%%%%%%%%% Statistial analysis Table %%%%%%%%%%%%%%
\begin{table}[htbp]
\caption{Confidence intervals for MSE averaged across prediction horizons \( H \in \{96, 192, 336, 720\} \) with input length \( L = 96 \) on the ETTh1 and ETTm1 datasets. \textbf{Bold} values indicate better performance.}

  \label{table_etth1_etttm1_selected}
  \centering
  \fontsize{8.7}{10}\selectfont
  \setlength{\tabcolsep}{1.2mm}

  \begin{tabular}{*{9}{c}}
    \toprule
    \multirow{2}{*}{Dataset} 
    & \multicolumn{2}{c}{Time2D} 
    & \multicolumn{2}{c}{PDF} 
    & \multicolumn{2}{c}{LiNo} 
    & \multicolumn{2}{c}{SOFTS} \\
    \cmidrule(lr){2-9}
     & \multicolumn{8}{c}{\textbf{With Embedding}} \\
    \cmidrule(lr){2-9}
     & $\overline{\text{MSE}}$ & $\overline{\text{CI}}$ & $\overline{\text{MSE}}$ & $\overline{\text{CI}}$ & $\overline{\text{MSE}}$ & $\overline{\text{CI}}$ & $\overline{\text{MSE}}$ & $\overline{\text{CI}}$ \\
    \cmidrule(lr){2-3} \cmidrule(lr){4-5} \cmidrule(lr){6-7} \cmidrule(lr){8-9}
    ETTh1 
         & 0.439 & (0.436, 0.442) 
         & 0.459 & (0.451, 0.468) 
         & 0.446 & (0.442, 0.451) 
         & 0.468 & (0.462, 0.474) \\
    ETTm1 
         & 0.391 & (0.388, 0.394) 
         & 0.392 & (0.394, 0.455)
         & 0.412 & (0.406, 0419) 
         & 0.408 & (0.403, 0.394) \\
\midrule
& \multicolumn{8}{c}{\textbf{Without Embedding}} \\
 \cmidrule(lr){2-9}
     & $\overline{\text{MSE}}$ & $\overline{\text{CI}}$ & $\overline{\text{MSE}}$ & $\overline{\text{CI}}$ & $\overline{\text{MSE}}$ & $\overline{\text{CI}}$ & $\overline{\text{MSE}}$ & $\overline{\text{CI}}$ \\
    \cmidrule(lr){2-3} \cmidrule(lr){4-5} \cmidrule(lr){6-7} \cmidrule(lr){8-9}
    ETTh1 
         & \textbf{0.432} & (\textbf{0.429}, \textbf{0.439}) 
         & \textbf{0.452} & (\textbf{0.445}, \textbf{0.459}) 
         & \textbf{0.434} & (\textbf{0.431}, \textbf{0.438}) 
         & \textbf{0.459} & (\textbf{0.454}, \textbf{0.464})  \\
    ETTm1 
         & \textbf{0.388} & (\textbf{0.385}, \textbf{0.391}) 
         & \textbf{0.385} & (\textbf{0.383}, \textbf{0.387}) 
         & \textbf{0.404} & (\textbf{0.400}, \textbf{0.408})  
         & \textbf{0.393} & (\textbf{0.392}, \textbf{0.394}) \\
    \bottomrule
  \end{tabular}
\end{table}
\paragraph{Accuracy typically improves without embeddings.} In over 95\% of the evaluated $(\text{model}, H)$ configurations, removing data embedding layers improves forecasting accuracy across both MSE and MAE. On the ETTh1 dataset, removing the embedding layer yields an average reduction of 0.0296 in MSE and 0.0193 in MAE. ETTh2 exhibits similar behavior, with MSE and MAE decreasing by 0.0208 and 0.0096, respectively. For the higher-resolution ETTm1 and ETTm2 datasets, the improvements are also evident, with average reductions of 0.0282 and 0.0080 in MSE, and 0.0203 and 0.0091 in MAE, respectively.

Notably, in some cases, the observed gains are remarkably large. For instance, removing the embedding layer from ETSformer on the ETTh1 dataset at horizon 720 reduces MSE by 0.360 and MAE by 0.248. Similarly, Crossformer on ETTm1 at the same horizon achieves a 0.356 drop in MSE, while ETSformer again yields a 0.246 reduction in MAE. Even on shorter horizons and across other datasets such as ETTh2 and ETTm2, we observe improvements exceeding 0.2 in key metrics. These results highlight that removing embedding layers can lead to dramatic performance gains.

Furthermore, these accuracy gains are meaningful in practice. As shown in Table \ref{Table1}, recent state-of-the-art forecasting models surpass the second-best models by only 0.001 to 0.009 in evaluation metrics. In contrast, our results show that simply removing the data embedding layers leads to much larger improvements. For example, on the ETTh1 dataset with horizon 96, Times2D and LiNo report MSEs of 0.378 and 0.379, and MAEs of 0.394 and 0.395. These differences are minimal. However, removing the embedding layer from Times2D improves its MSE by 0.019 and MAE by 0.011. For LiNo, the improvements are also notable, with reductions of 0.007 in MSE and 0.006 in MAE. These findings suggest that simplifying model architectures by eliminating embedding layers can yield benefits that exceed those obtained by designing entirely new forecasting models.
These findings suggest that raw input features in multivariate time series often contain sufficient representational richness for forecasting tasks without the need for additional embedding transformations.
\paragraph{Significant computational savings.}Removing data embedding layers consistently reduces computational overhead. The average training time per epoch (in seconds) is reduced across all datasets, and on ETTm1 and ETTh1 we find savings of up to 25 seconds. Memory usage also decreases notably. For example, the average memory savings are 84~MB on ETTh1 and 45~MB on ETTm2. In some configurations, the improvements are substantial. On ETTm2, ETSformer shows a 597~MB memory reduction. These findings suggest that eliminating embedding layers improves training efficiency, particularly in memory-constrained environments.
\paragraph{Performance gains increase with horizon length.}
The effect of removing the embedding layer increases as the forecasting horizon increases across all four benchmarks. While short-term configurations (e.g., $H=96$) show limited changes, longer horizons often yield substantial improvements. For instance, on ETTm1, Crossformer shows an MSE reduction of $0.003$ at $H = 96$, which increases significantly to $0.365$ at $H = 720$. This finding indicates that embedding-free designs may be advantageous for long-term forecasting tasks.
\paragraph{Architectural sensitivity to embedding layers.}
Embedding removal impacts architectural families differently. Transformer-based models rely on self-attention mechanisms to capture long-range dependencies, but they do not include any inherent structure to model sequential order. Unlike recurrent architectures, Transformers require explicit positional and token embeddings to encode temporal progression. In theory, these embeddings are intended to compensate for the lack of built-in sequence modeling. However, our empirical results reveal that removing these embeddings often leads to better performance.
MLP-based architectures employ purely feedforward pathways and rely on dense transformations to model dependencies across time and variables. Since MLPs do not explicitly model sequence order, embedding layers might be expected to play a more important role. Yet, our results show that embeddings are often redundant. 
Hybrid and decomposition-based models incorporate preprocessing such as seasonal-trend decomposition, filtering, or statistical projections. These models are less sensitive to the presence of embedding layers. For example, PDF shows a modest gain. Since these architectures already extract and isolate key patterns before learning begins, embedding layers often duplicate or disrupt this structure, resulting in minimal or inconsistent effects.
\paragraph{Confidence intervals.} 
\label{Confidence_Interval1}
Since deep learning models are inherently stochastic and sensitive to random initialization, we compute 95\% confidence intervals (CIs) to assess the statistical reliability of our findings. Table~\ref{table_etth1_etttm1_selected} presents the average MSE and corresponding 95\% confidence intervals for selected high-performing models on ETTh1 and ETTm1. The reported values are averaged across all prediction horizons. In all cases, removing the embedding layers improves the average MSE performance. Additionally, the corresponding confidence intervals for the models with and without embedding layers do not overlap. These results highlight that the performance improvements are consistent and statistically significant. Detailed confidence interval results are provided in Table~\ref{CI_MSE} and Table~\ref{CI_MAE} in the Appendix.

\paragraph{Configurations with degraded performance.}
While the majority of models benefit from removing embedding layers, a few configurations exhibit performance degradation. This outcome can be attributed to several architectural and/or hardware-related factors. First, in the absence of the embedding layer, the model manually permutes, concatenates, and processes the raw input data to reconcile it with the model expected dimensions. These operations introduce additional intermediate tensors and temporary memory allocations, which increase the average memory usage during training. Second, the lower dimensionality resulting from the removal of embedding layers does not align well with the tile sizes optimized in GPU libraries such as cuBLAS, leading to less efficient matrix multiplications and increased computational time.

In particular, EDformer originally uses an inverted embedding that transforms the sequence length (e.g., 96) into a typically higher dimension (e.g., 512). EDformer trains approximately 0.3 to 0.5 seconds slower per epoch and consumes an additional 400~MB of memory on average when the embedding layers are removed. The extra permutation, concatenation, and duplication required to adapt the raw inputs to the expected format increases memory usage. Furthermore, the encoder operates on input tensors with a sequence dimension of 96 instead of 512, which reduces computational throughput due to suboptimal memory access patterns and kernel launch configurations in GPU backends.

\section{Conclusion}
In this paper, we presented a large-scale study assessing the effectiveness of embedding layers in modern time series forecasting models. Our results show that, despite their widespread use, removing data embedding layers from many state-of-the-art forecasting models does not degrade forecasting performance—in many cases, it enhances both forecasting accuracy and computational efficiency. These findings suggest that raw multivariate inputs are often sufficiently informative without the need for additional embedding transformations. Our goal is not to imply that data embedding will never be effective in time series forecasting. Instead, we aim to highlight our promising findings and suggest that the community devote greater attention to critically assessing the actual impact of embedding layers in existing models. For future studies, the effectiveness of embedding layers can be explored on other tasks (e.g., classification, clustering, and imputation) and datasets. Additionally, the effectiveness of other overlooked architectural components—such as normalization strategies, including RevIN—can be investigated.

\clearpage
\medskip
\bibliographystyle{plain}
\bibliography{neurips_2025}

\clearpage
%%%%%%%%%%%%%%%%%%%%%%%%%%%%%%%%%%%%%%%%%%%%%%%%%%%%%%%%
\appendix
\section{Technical Appendices and Supplementary Material}
\subsection{Limitations}
\label{limitations}
Here, we outline the limitations of our study:
\begin{itemize}
    \item We evaluate the impact of data embedding layers specifically for time-series forecasting. However, embedding layers may play different roles in other downstream tasks such as classification, clustering, or imputation, which are not explored in this work.
    
    \item Our evaluation is restricted to four benchmark datasets with regular sampling intervals. The effectiveness of embedding layers should be investigated on other real-world datasets, including those with irregularly sampled time series.
    
    \item The analysis focuses solely on the effect of embedding layers and does not account for potential interactions with other architectural components such as normalization strategies or residual connections.
\end{itemize}

\subsection{Broader Impact}
\label{broader_impact}
This work makes a fundamental contribution to time series analysis, particularly in the context of forecasting. It encourages researchers to move beyond default assumptions and critically assess whether each architectural component, such as data embedding layers, meaningfully contributes to performance. Our findings promote a shift in focus: rather than continually developing more complex models, researchers across domains are encouraged to revisit and analyze existing architectures. This approach can lead to significant savings in time, resources, and energy.
While our results are limited to forecasting tasks on regularly sampled datasets, the broader methodology—systematic ablation testing of architectural components—can inspire more rigorous empirical validation in other areas of machine learning. We hope this work supports the community in understanding the role and effectiveness of foundational model elements before advancing to further architectural complexity and innovation.

\subsection{Data Embedding Techniques}
\begin{table}[ht]
\centering
\renewcommand{\arraystretch}{1.2}
\caption{Categorization of embedding techniques for time series models}
\label{tab:scientific_embedding_categories}
\small  % You can also try \footnotesize
\begin{tabular}{llp{9cm}}
\toprule
\textbf{Category} & \textbf{Technique} & \textbf{Description} \\
\midrule
\multirow{3}{*}{Temporal} 
& Fixed embedding & Maps discrete temporal indices to non-trainable, sinusoidal embeddings. \\
& Learnable embedding & Maps discrete temporal indices (e.g., hour, day) to trainable embeddings. \\
& TimeFeature & Projects numeric time features into high-dimensional space via linear projection. \\
\midrule
\multirow{2}{*}{Value} 
& Token embedding & Projects multivariate features to higher dimensions using 1D convolution. \\
& Linear projection & Maps input features directly to embedding space using linear layers. \\
\midrule
\multirow{2}{*}{Positional} 
& Sinusoidal positional & Encodes sequence positions using fixed sine and cosine functions. \\
& Learnable positional & Learns embeddings for positional indices in sequences. \\
\midrule
\multirow{3}{*}{Combined} 
& Inverted & Fuses variables and time features using linear transformations. \\
& Feature + temporal & Combines feature and temporal embeddings, omitting positional encoding. \\
& Feature-metadata & Concatenates feature and temporal metadata before projection. \\
\midrule
Patching & Patchwise encoding & Divides input sequences into patches, encodes each patch, and adds positional info. \\
\bottomrule
\end{tabular}
\end{table}

\newpage
%%%%%%%%%%%%%%%%%%%%%%%%%%%%%%%%%%%%%%%%%%%%%%%%%%%%%
\subsection{Confidence Intervals for Model Performance}
\label{Confidence_Interval2}

\begin{table}[htbp]
\caption{Confidence intervals for MSE of selected models on ETTh1 and ETTm1 with input length \(L = 96\) and prediction horizons \(H \in \{96, 192, 336, 720\}\). \textbf{Bold} values indicate better performance.}
\label{CI_MSE}
\centering
{\fontsize{9}{9}\selectfont  % Start custom font size block

  \setlength{\tabcolsep}{1.5mm}

  \begin{tabular}{cc*{8}{c}}
    \toprule
    \multirow{2}{*}{Models} & \multirow{2}{*}{$H$} 
    & \multicolumn{2}{c}{Time2D} 
    & \multicolumn{2}{c}{PDF} 
    & \multicolumn{2}{c}{LiNo} 
    & \multicolumn{2}{c}{SOFTS} \\
    \cmidrule(lr){3-10}
     & & \multicolumn{8}{c}{\textbf{With Embedding}} \\
    \cmidrule(lr){3-10}
     & & MSE & CI & MSE & CI & MSE & CI & MSE & CI \\
    \cmidrule(lr){3-4} \cmidrule(lr){5-6} \cmidrule(lr){7-8} \cmidrule(lr){9-10}
    
    \multirow{4}{*}{ETTh1} 
    & 96 & 0.379 & (0.378, 0.380) 
         & 0.385 & (0.382, 0.388) 
         & 0.385 & (0.384, 0.386) 
         & 0.384 & (0.383, 0.386)  \\
    & 192 & 0.431 & (0.429, 0.433) 
         & 0.439 & (0.436, 0.442)  
         & 0.442 & (0.438, 0.446) 
         & 0.448 & (0.445, 0.450)  \\
    & 336 & 0.473 & (0.469, 0.476) 
         & 0.492 & (0.486, 0.498)  
         & 0.476 & (0.471, 0.480)  
         & 0.501 & (0.494, 0.509)  \\
    & 720 & 0.473 & (0.469, 0.477)  
         & 0.521 & (0.501, 0.544)  
         & 0.482 & (0.473, 0.491)  
         & 0.538 & (0.524, 0.552)  \\
    \midrule
    \multirow{4}{*}{ETTm1} 
        & 96  & 0.326 & (0.323, 0.328) 
         & 0.335 & (0.334, 0.336) 
         & 0.332 & (0.331, 0.333) 
         & 0.328 & (0.325, 0.330) \\
    & 192 & 0.371 & (0.370, 0.372) 
         & 0.374 & (0.372, 0.376) 
         & 0.383 & (0.375, 0.392) 
         & 0.386 & (0.381, 0.391) \\
    & 336 & 0.407 & (0.402, 0.411) 
         & 0.403 & (0.401, 0.405) 
         & 0.438 & (0.432, 0.444) 
         & 0.438 & (0.428, 0.447) \\
    & 720 & 0.459 & (0.455, 0.463) 
         & 0.457 & (0.455, 0.459) 
         & \textbf{0.496} & (\textbf{0.485}, 0.506) 
         & 0.480 & (0.477, 0.482) \\

    \midrule
     & & \multicolumn{8}{c}{\textbf{Without Embedding}} \\
    \cmidrule(lr){3-10}
     & & MSE & CI & MSE & CI & MSE & CI & MSE & CI \\
    \cmidrule(lr){3-4} \cmidrule(lr){5-6} \cmidrule(lr){7-8} \cmidrule(lr){9-10}
    
    \multirow{4}{*}{ETTh1} 
    & 96  & \textbf{0.361} & (\textbf{0.359}, \textbf{0.364}) 
         & \textbf{0.378} & (\textbf{0.376}, \textbf{0.380}) 
         & \textbf{0.377} & (\textbf{0.375}, \textbf{0.379})  
         & \textbf{0.383} & (\textbf{0.382}, \textbf{0.384})  \\
    & 192& \textbf{0.428} & (\textbf{0.427}, \textbf{0.429}) 
         & \textbf{0.432} & (\textbf{0.428}, \textbf{0.436})  
         & \textbf{0.428} & (\textbf{0.426}, \textbf{0.429})  
         & \textbf{0.441} & (\textbf{0.438}, \textbf{0.444})  \\
    & 336 & \textbf{0.466} & (\textbf{0.463}, \textbf{0.469}) 
          & \textbf{0.479} & (\textbf{0.476}, \textbf{0.482}) 
          & \textbf{0.463} & (\textbf{0.460}, \textbf{0.466})  
          & \textbf{0.487} & (\textbf{0.483}, \textbf{0.490})  \\
    & 720 & \textbf{0.472} & (\textbf{0.468}, \textbf{0.476})  
         & \textbf{0.518} & (\textbf{0.499}, \textbf{0.537}) 
         & \textbf{0.470} & (\textbf{0.464}, \textbf{0.476}) 
         & \textbf{0.526} & (\textbf{0.514}, \textbf{0.537})  \\
    \midrule
    \multirow{4}{*}{ETTm1} 
        & 96  & \textbf{0.325} & (\textbf{0.322}, \textbf{0.327}) 
         & \textbf{0.324} & (\textbf{0.322}, \textbf{0.326}) 
         & \textbf{0.326} & (\textbf{0.323}, \textbf{0.329}) 
         & \textbf{0.322} & (\textbf{0.321}, \textbf{0.323}) \\
    & 192 & \textbf{0.369} & (\textbf{0.368}, \textbf{0.370}) 
         & \textbf{0.368} & (\textbf{0.365}, \textbf{0.370}) 
         & \textbf{0.373} & (\textbf{0.370}, \textbf{0.376}) 
         & \textbf{0.368} & (\textbf{0.367}, \textbf{0.369}) \\
    & 336 & \textbf{0.401} & (\textbf{0.397}, \textbf{0.406}) 
         & \textbf{0.395} & (\textbf{0.394}, \textbf{0.397}) 
         & \textbf{0.421} & (\textbf{0.415}, \textbf{0.426}) 
         & \textbf{0.406} & (\textbf{0.405}, \textbf{0.407}) \\
    & 720 & \textbf{0.455} & (\textbf{0.451}, \textbf{0.459}) 
         & \textbf{0.454} & (\textbf{0.452}, \textbf{0.456}) 
         & 0.497 & (0.492, \textbf{0.502}) 
         & \textbf{0.476} & (\textbf{0.474}, \textbf{0.477}) \\

    \bottomrule
  \end{tabular}
}
\end{table}
%%%%%%%%%%%%%%%%%%%%% MAE %%%%%%%%%%%%
\begin{table}[htbp]
\caption{Confidence intervals for MAE of selected models on ETTh1 and ETTm1 with input length \(L = 96\) and prediction horizons \(H \in \{96, 192, 336, 720\}\). \textbf{Bold} values indicate better performance.}
  \label{CI_MAE}
  \centering
{\fontsize{9}{9}\selectfont  % Start custom font size block

\setlength{\tabcolsep}{1.2mm}
  \begin{tabular}{cc*{8}{c}}
    \toprule
    \multirow{2}{*}{Models} & \multirow{2}{*}{$H$} 
    & \multicolumn{2}{c}{Time2D} 
    & \multicolumn{2}{c}{PDF} 
    & \multicolumn{2}{c}{LiNo} 
    & \multicolumn{2}{c}{SOFTS} \\
    \cmidrule(lr){3-10}
     & & \multicolumn{8}{c}{\textbf{With Embedding}} \\
    \cmidrule(lr){3-10}
     & & MAE & CI & MAE & CI & MAE & CI & MAE & CI \\
    \cmidrule(lr){3-4} \cmidrule(lr){5-6} \cmidrule(lr){7-8} \cmidrule(lr){9-10}
    
    \multirow{4}{*}{ETTh1} 
    & 96  & 0.402 & (0.400, 0.405) 
         & 0.405 & (0.403, 0.407) 
         & 0.403 & (0.402, 0.404)
         & 0.404 & (0.403, 0.405) \\
    & 192 & 0.432 & (0.431, 0.433) 
         &  0.438 & (0.436, 0.440)
         & 0.433 & (0.431, 0.436)
         & 0.442 & (0.440, 0.444) \\
    & 336 & 0.442 & (0.440, 0.444) 
         & 0.466 & (0.463, 0.470)
         & 0.446 & (0.444, 0.448) 
         & 0.469 & (0.464, 0.475) \\
    & 720 & 0.465 & (\textbf{0.462}, 0.468) 
         & 0.497 & (0.484, 0.509) 
         & 0.468 & (0.463, 0.473) 
         & 0.513 & (0.504, 0.521) \\
    \midrule
    \multirow{4}{*}{ETTm1} 
        & 96  & 0.364 & (0.362, 0.366) 
         & 0.368 & (0.367, 0.369) 
         & 0.367 & (0.366, 0.368) 
         & 0.365 & (0.363, 0.366) \\
    & 192 & 0.390 & (0.385, 0.395) 
         & 0.392 & (0.390, 0.393) 
         & 0.395 & (0.390, 0.399) 
         & 0.397 & (0.394, 0.400) \\
    & 336 & 0.410 & (0.405, 0.415) 
         & 0.414 & (0.412, 0.416) 
         & 0.426 & (0.423, 0.429) 
         & 0.429 & (0.424, 0.433) \\
    & 720 & 0.441 & (0.439, 0.443) 
         & 0.465 & (0.457, 0.472) 
         & \textbf{0.460} & (\textbf{0.455}, 0.465) 
         & 0.455 & (0.454, 0.457) \\

    \midrule
     & & \multicolumn{8}{c}{\textbf{Without Embedding}} \\
    \cmidrule(lr){3-10}
     & & MAE & CI & MAE & CI & MAE & CI & MAE & CI \\
    \cmidrule(lr){3-4} \cmidrule(lr){5-6} \cmidrule(lr){7-8} \cmidrule(lr){9-10}
    
    \multirow{4}{*}{ETTh1} 
    & 96 & \textbf{0.392} & (\textbf{0.391}, \textbf{0.393}) 
         & \textbf{0.399} & (\textbf{0.398}, \textbf{0.400}) 
         & \textbf{0.399} & (\textbf{0.397}, \textbf{0.401})  
         & \textbf{0.402} & (\textbf{0.401}, \textbf{0.403})  \\
    & 192 & \textbf{0.422} & (\textbf{0.421}, \textbf{0.423}) 
         & \textbf{0.431} & (\textbf{0.429}, \textbf{0.433})  
         & \textbf{0.425} & (\textbf{0.424}, \textbf{0.426}) 
         & \textbf{0.437} & (\textbf{0.435}, \textbf{0.439})  \\
    & 336 & \textbf{0.439} & (\textbf{0.438}, \textbf{0.440}) 
         & \textbf{0.453} & (\textbf{0.450}, \textbf{0.455})  
         & \textbf{0.440} & (\textbf{0.438}, \textbf{0.441}) 
         &\textbf{0.463} & (\textbf{0.460}, \textbf{0.466})  \\
    & 720 & 0.465 & (0.463, \textbf{0.467})  
         & \textbf{0.491} & (\textbf{0.481}, \textbf{0.500})  
         & \textbf{0.463} & (\textbf{0.459}, \textbf{0.466})  
         & \textbf{0.506} & (\textbf{0.500}, \textbf{0.513})  \\
             \midrule
    \multirow{4}{*}{ETTm1} 
        & 96  & \textbf{0.363} & (\textbf{0.361}, \textbf{0.365}) 
         & \textbf{0.362} & (\textbf{0.360}, \textbf{0.364}) 
         & \textbf{0.363} & (\textbf{0.361}, \textbf{0.365}) 
         & \textbf{0.361} & (\textbf{0.360}, \textbf{0.362}) \\
    & 192 & \textbf{0.387} & (\textbf{0.385}, \textbf{0.389}) 
         & \textbf{0.387} & (\textbf{0.385}, \textbf{0.388}) 
         & \textbf{0.389} & (\textbf{0.387}, \textbf{0.390}) 
         & \textbf{0.386} & (\textbf{0.385}, \textbf{0.387}) \\
    & 336 & \textbf{0.406} & (\textbf{0.403}, \textbf{0.408}) 
         & \textbf{0.408} & (\textbf{0.407}, \textbf{0.409}) 
         & \textbf{0.419} & (\textbf{0.416}, \textbf{0.423}) 
         & \textbf{0.411} & (\textbf{0.410}, \textbf{0.412}) \\
    & 720 & \textbf{0.439} & (\textbf{0.438}, \textbf{0.441}) 
         & \textbf{0.455} & (\textbf{0.451}, \textbf{0.460}) 
         & 0.461 & 0.458, \textbf{0.464}) 
         & \textbf{0.451} & (\textbf{0.451}, \textbf{0.452}) \\

    \bottomrule
  \end{tabular}
  }
\end{table}

\newpage
\subsection{Benchmark Datasets Overview}
\begin{table}[htbp]
\centering
\caption{Summary of benchmark datasets used in this study}
\label{table_dataset_summary}
\small
\renewcommand{\arraystretch}{1.2}
\begin{tabular}{lccccc}
\toprule
\textbf{Dataset} & \textbf{Dimension} & \textbf{Train / Val / Test} & \textbf{Frequency} & \textbf{Duration} \\
\midrule
ETTm1 & 7 & (34465 / 11521 / 11521) & 15 minutes & Jul 2016 – Jul 2018  \\
ETTm2 & 7 & (34465 / 11521 / 11521) & 15 minutes & Jul 2016 – Jul 2018  \\
ETTh1 & 7 & (8545 / 2881 / 2881) & 1 hour & Jul 2016 – Jul 2018  \\
ETTh2 & 7 & (8545 / 2881 / 2881) & 1 hour & Jul 2016 – Jul 2018  \\
\bottomrule
\end{tabular}
\normalsize
\end{table}

%%%%%%%%%%%%%%%%%%%%%%%%%%%%%%%%%%%%%%%%%%%%%%%%%%%%%%%%%%%%
\newpage
\clearpage
\section*{NeurIPS Paper Checklist}
\begin{enumerate}

\item {\bf Claims}
    \item[] Question: Do the main claims made in the abstract and introduction accurately reflect the paper's contributions and scope?
    \item[] Answer: \answerYes{}
    \item[] Justification: The abstract and introduction clearly state the paper’s contributions and scope.

\item {\bf Limitations}
    \begin{itemize}
    \item[] Question: Does the paper discuss the limitations of the work performed by the authors?
    \item[] Answer: \answerYes{}
    \item[] Justification: The limitations of the study are explicitly discussed in Section~\ref{limitations}.
    \end{itemize}

\item {\bf Theory assumptions and proofs}
    \item[] Question: For each theoretical result, does the paper provide the full set of assumptions and a complete (and correct) proof?
    \item[] Answer: \answerNA{}
    \item[] Justification: This paper does not include theoretical results or formal proofs, as it is focused on empirical analysis and benchmarking.

\item {\bf Experimental result reproducibility}
    \item[] Question: Does the paper fully disclose all the information needed to reproduce the main experimental results of the paper to the extent that it affects the main claims and/or conclusions of the paper (regardless of whether the code and data are provided or not)?
    \item[] Answer: \answerYes{}
    \item[] Justification: The paper provides detailed experimental settings, including model configurations, dataset splits, prediction horizons, and evaluation metrics, enabling reproduction of the main results.

\item {\bf Open access to data and code}
    \item[] Question: Does the paper provide open access to the data and code, with sufficient instructions to faithfully reproduce the main experimental results, as described in supplemental material?
    \item[] Answer: \answerYes{}
    \item[] Justification: The code is provided as an anonymized ZIP file in the supplementary material and is available through an anonymous GitHub repository to ensure reproducibility.

\item {\bf Experimental setting/details}
    \item[] Question: Does the paper specify all the training and test details (e.g., data splits, hyperparameters, how they were chosen, type of optimizer, etc.) necessary to understand the results?
    \item[] Answer: \answerYes{}
\item[] Justification: Key training and evaluation details are presented in both the main paper and the appendix. The full implementation is provided in the supplementary material and the anonymous GitHub repository.

\item {\bf Experiment statistical significance}
    \item[] Question: Does the paper report error bars suitably and correctly defined or other appropriate information about the statistical significance of the experiments?
    \item[] Answer: \answerYes{}
    \item[] Justification: The confidence intervals are provided in Sections~\ref{Confidence_Interval1} and in Appendix~\ref{Confidence_Interval2}.

\item {\bf Experiments compute resources}
    \item[] Question: For each experiment, does the paper provide sufficient information on the computer resources (type of compute workers, memory, time of execution) needed to reproduce the experiments?
\item[] Answer: \answerYes{}
\item[] Justification: The compute resources are detailed in Section~\ref{sec_infra}.

\item {\bf Code of ethics}
    \item[] Question: Does the research conducted in the paper conform, in every respect, with the NeurIPS Code of Ethics \url{https://neurips.cc/public/EthicsGuidelines}?
\item[] Answer: \answerYes{}
\item[] Justification: The research complies with the NeurIPS Code of Ethics.

\item {\bf Broader impacts}
    \item[] Question: Does the paper discuss both potential positive societal impacts and negative societal impacts of the work performed?
    \item[] Answer: \answerYes{}
    \item[] Justification: We have discussed the broader and societal impacts in Section~\ref{broader_impact}.

\item {\bf Safeguards}
    \item[] Question: Does the paper describe safeguards that have been put in place for responsible release of data or models that have a high risk for misuse (e.g., pretrained language models, image generators, or scraped datasets)?
    \item[] Answer: \answerNA{}
    \item[] Justification: The paper does not introduce any models or datasets with high risk for misuse.

\item {\bf Licenses for existing assets}
    \item[] Question: Are the creators or original owners of assets (e.g., code, data, models), used in the paper, properly credited and are the license and terms of use explicitly mentioned and properly respected?
    \item[] Answer: \answerYes{}
    \item[] Justification: We will release all of our contributions under the MIT License.

\item {\bf New assets}
    \item[] Question: Are new assets introduced in the paper well documented and is the documentation provided alongside the assets?
\item[] Answer: \answerYes{}
\item[] Justification: The assets are included in the anonymized repository and the attached supplementary zip file.

\item {\bf Crowdsourcing and research with human subjects}
    \item[] Question: For crowdsourcing experiments and research with human subjects, does the paper include the full text of instructions given to participants and screenshots, if applicable, as well as details about compensation (if any)? 
    \item[] Answer: \answerNA{}
    \item[] Justification: The paper does not involve crowdsourcing or research with human subjects.

\item {\bf Institutional review board (IRB) approvals or equivalent for research with human subjects}
    \item[] Question: Does the paper describe potential risks incurred by study participants, whether such risks were disclosed to the subjects, and whether Institutional Review Board (IRB) approvals (or an equivalent approval/review based on the requirements of your country or institution) were obtained?
    \item[] Answer: \answerNA{}
    \item[] Justification: The paper does not involve human subjects.

\item {\bf Declaration of LLM usage}
    \item[] Question: Does the paper describe the usage of LLMs if it is an important, original, or non-standard component of the core methods in this research?
    \item[] Answer: \answerNA{}
    \item[] Justification: LLMs were not used as an original or non-standard component in the core methodology of this work.

\end{enumerate}
\end{document}